\documentclass[11pt]{article}
\usepackage{coling2018}
\usepackage{times}
\usepackage{latexsym}
\usepackage{anysize}
\usepackage{graphicx}
\usepackage{subfig}
\usepackage{mathtools}
\usepackage{amsmath}
\usepackage{bm}
\usepackage{enumerate}
\usepackage{multirow}
\usepackage{algpseudocode}
\usepackage{algorithmicx}
\usepackage{caption}
\usepackage{float}
\usepackage{url}
\usepackage{array}
\usepackage{tabulary}

\title{Author Profiling for Hate Speech Detection}

\author{
  Pushkar Mishra\\
  Dept. of CS and Technology\\
  University of Cambridge\\
  United Kingdom\\
  {\tt pm576@cl.cam.ac.uk} \\\And
  Marco Del Tredici\\
  ILLC\\
  University of Amsterdam\\
  The Netherlands\\
  {\tt m.deltredici@uva.nl} \\\AND
  Helen Yannakoudakis\\
  Dept. of CS and Technology\\
  The ALTA Institute \\
  University of Cambridge\\
  United Kingdom\\
  {\tt hy260@cl.cam.ac.uk} \\\And
  Ekaterina Shutova\\
  ILLC\\
  University of Amsterdam\\
  The Netherlands\\
  {\tt e.shutova@uva.nl}\\
}

\date{}

\begin{document}
\maketitle
\begin{abstract}
The rapid growth of social media in recent years has fed into some highly undesirable phenomena such as proliferation of abusive and offensive language on the Internet. Previous research suggests that such hateful content tends to come from users who share a set of common stereotypes and form communities around them.
The current state-of-the-art approaches to hate speech detection are oblivious to user and community information and rely entirely on textual (i.e., lexical and semantic) cues. 
In this paper, we propose a novel approach to this problem that incorporates community-based profiling features of Twitter users. Experimenting with a dataset of $16k$ tweets, we show that our methods significantly outperform the current state of the art in hate speech detection. Further, we conduct a qualitative analysis of model characteristics. We release our code, pre-trained models and all the resources used in the public domain.
\end{abstract}

\section{Introduction}
\blfootnote{\hspace{-6mm}This work is licensed under a Creative Commons Attribution 4.0 International License.\\License details: \url{http://creativecommons.org/licenses/by/4.0/}.}
Hate speech, a term used to collectively refer to offensive language, racist comments, sexist remarks, etc., is omnipresent in social media. Users on social media platforms are at risk of being exposed to content that may not only be degrading but also harmful to their mental health in the long term. \textit{Pew Research Center} highlighted the gravity of the situation via a recently released report \cite{pew}. As per the report, 40\% of adult Internet users have personally experienced harassment online, and 60\% have witnessed the use of offensive names and expletives. Expectedly, the majority (66\%) of those who have personally faced harassment have had their most recent incident occur on a social networking website or app. While most of these websites and apps provide ways of flagging offensive and hateful content, only 8.8\% of the victims have actually considered using such provisions. These statistics suggest that passive or manual techniques for curbing propagation of hateful content (such as flagging) are neither effective nor easily scalable \cite{Pavlopoulos:17}. Consequently, the efforts to automate the detection and moderation of such content have been gaining popularity in natural language processing (\textsc{nlp}) \cite{c53cecce142c48628b3883d13155261c,Wulczyn:2017:EMP:3038912.3052591}.

Several approaches to hate speech detection demonstrate the effectiveness of character-level bag-of-words features in a supervised classification setting \cite{Djuric:2015:HSD:2740908.2742760,Nobata:2016:ALD:2872427.2883062,davidson}. More recent approaches, and currently the best performing ones, utilize recurrent neural networks (\textsc{rnn}s) to transform content into dense low-dimensional semantic representations that are then used for classification \cite{Pavlopoulos:17,Badjatiya:17}. All of these approaches rely solely on lexical and semantic features of the text they are applied to. Waseem and Hovy \shortcite{c53cecce142c48628b3883d13155261c} adopted a more user-centric approach based on the idea that perpetrators of hate speech are usually segregated into small demographic groups; they went on to show that gender information of \textit{authors} (i.e., users who have posted content) is a helpful indicator. However, Waseem and Hovy focused only on coarse demographic features of the users, disregarding information about their communication with others. But previous research suggests that users who subscribe to particular stereotypes that promote hate speech tend to form communities online. For example, Zook \shortcite{zook} mapped the locations of racist tweets in response to President Obama's re-election to show that such tweets were not uniformly distributed across the United States but formed clusters instead. In this paper, we present the first approach to hate speech detection that leverages author profiling information based on properties of the authors' social network and investigate its effectiveness.

Author profiling has emerged as a powerful tool for NLP applications, leading to substantial performance improvements in several downstream tasks, such as text classification, sentiment analysis and author attribute identification \cite{hovy2015demographic,eisenstein2015written,yang2017overcoming}. 
The relevance of information gained from it is best explained by the idea of \textit{homophily}, i.e., the phenomenon that people, both in real life as well as on the Internet, tend to associate more with those who appear similar. Here, similarity can be defined along various axes, e.g., location, age, language, etc. The strength of author profiling lies in that if we have information about members of a community $c$ defined by some similarity criterion, and we know that the person $p$ belongs to $c$, we can infer information about $p$. This concept has a straightforward application to our task: knowing that members of a particular community are prone to creating hateful content, and knowing that the author \textit{p} is connected to this community, we can leverage information beyond linguistic cues and more accurately predict the use of hateful/non-hateful language from $p$. The questions that we seek to address here are: are some authors, and the respective communities that they belong to, more hateful than the others? And can such information be effectively utilized to improve the performance of automated hate speech detection methods?

In this paper, we answer these questions and develop novel methods that take into account community-based profiling features of authors when examining their tweets for hate speech. Experimenting with a dataset of $16k$ tweets, we show that the addition of such profiling features to the current state-of-the-art methods for hate speech detection significantly enhances their performance. We also release our code (including code that replicates previous work), pre-trained models and the resources we used in the public domain. 


\section{Related Work}
\subsection{Hate speech detection}
Amongst the first ones to apply supervised learning to the task of hate speech detection were Yin et al. \shortcite{Yin09detectionof} who used a linear \textsc{svm} classifier to identify posts containing harassment based on local (e.g., n-grams), contextual (e.g., similarity of a post to its neighboring posts) and sentiment-based (e.g., presence of expletives) features. Their best results were with all of these features combined.

Djuric et al. \shortcite{Djuric:2015:HSD:2740908.2742760} experimented with comments extracted from the Yahoo Finance portal and showed that distributional representations of comments learned using \textit{paragraph2vec} \cite{DBLP:journals/corr/LeM14} outperform simpler bag-of-words (\textsc{bow}) representations in a supervised classification setting for hate speech detection. Nobata et al. \shortcite{Nobata:2016:ALD:2872427.2883062} improved upon the results of Djuric et al. by training their classifier on a combination of features drawn from four different categories: linguistic (e.g., count of insult words), syntactic (e.g., \textsc{pos} tags), distributional semantic (e.g., word and comment embeddings) and \textsc{bow}-based (word and characters n-grams). They reported that while the best results were obtained with all features combined, character n-grams contributed more to performance than all the other features.

Waseem and Hovy \shortcite{c53cecce142c48628b3883d13155261c} created and experimented with a dataset of racist, sexist and clean tweets. Utilizing a logistic regression (\textsc{lr}) classifier to distinguish amongst them, they found that character n-grams coupled with gender information of users formed the optimal feature set; on the other hand, geographic and word-length distribution features provided little to no improvement. Working with the same dataset, Badjatiya et al. \shortcite{Badjatiya:17} improved on their results by training a gradient-boosted decision tree (\textsc{gbdt}) classifier on averaged word embeddings learnt using a long short-term memory (\textsc{lstm}) network that they initialized with random embeddings.

Waseem \shortcite{zeerakW16-5618} sampled $7k$ more tweets in the same manner as Waseem and Hovy \shortcite{c53cecce142c48628b3883d13155261c}. They recruited expert and amateur annotators to annotate the tweets as \textit{racism}, \textit{sexism}, \textit{both} or \textit{neither} in order to study the influence of annotator knowledge on the task of hate speech detection. Combining this dataset with that of Waseem and Hovy \shortcite{c53cecce142c48628b3883d13155261c}, Park et al. \shortcite{W17-3006} explored the merits of a two-step classification process. They first used a \textsc{lr} classifier to separate hateful and non-hateful tweets, followed by another \textsc{lr} classifier to distinguish between racist and sexist ones. They showed that this setup had comparable performance to a one-step classification setup built with convolutional neural networks.
		
Davidson et al. \shortcite{davidson} created a dataset of about $25k$ tweets wherein each tweet was annotated as being \textit{racist}, \textit{offensive} or \textit{neither of the two}. They tested several multi-class classifiers with the aim of distinguishing clean tweets from racist and offensive tweets while simultaneously being able to separate the racist and offensive ones. Their best model was a \textsc{lr} classifier trained using \textsc{tf-idf} and \textsc{pos} n-gram features, as well as the count of hash tags and number of words.

Wulczyn et al. \shortcite{Wulczyn:2017:EMP:3038912.3052591} prepared three different datasets of comments collected from the English Wikipedia Talk page; one was annotated for personal attacks, another for toxicity and the third one for aggression. Their best performing model was a multi-layered perceptron (\textsc{mlp}) classifier trained on character n-gram features. Experimenting with the personal attack and toxicity datasets, Pavlopoulos et al. \shortcite{Pavlopoulos:17} improved the results of Wulczyn et al. by using a gated recurrent unit (\textsc{gru}) model to encode the comments into dense low-dimensional representations, followed by a \textsc{lr} layer to classify the comments based on those representations.

\subsection{Author profiling}
Author profiling has been leveraged in several ways for a variety of purposes in \textsc{nlp}. For instance, many studies have relied on demographic information of the authors. Amongst these are Hovy et al. \shortcite{hovy2015demographic} and Ebrahimi et al. \shortcite{ebrahimi2016personalized} who extracted age and gender-related information to achieve superior performance in a text classification task. Pavalanathan and Eisenstein \shortcite{pavalanathan2015confounds}, in their work, further showed the relevance of the same information to automatic text-based geo-location. Researching along the same lines, Johannsen et al. \shortcite{johannsen2015cross} and Mirkin et al. \shortcite{mirkin2015motivating} utilized demographic factors to improve syntactic parsing and machine translation respectively. 

While demographic information has proved to be relevant for a number of tasks, it presents a significant drawback: since this information is not always available for all authors in a social network, it is not particularly reliable. Consequently, of late, a new line of research has focused on creating representations of users in a social network by leveraging the information derived from the connections that they have with other users. In this case, node representations (where nodes represent the authors in the social network) are typically induced using neural architectures. Given the graph representing the social network, such methods create low-dimensional representations for each node, which are optimized to predict the nodes close to it in the network. This approach has the advantage of overcoming the absence of information that the previous approaches face. Among those that implement this idea are Yang et al. \shortcite{yang2016toward}, who used representations derived from a social graph to achieve better performance in entity linking tasks, and Chen and Ku \shortcite{chen2016utcnn}, who used them for stance classification.

A considerable amount of literature has also been devoted to sentiment analysis with representations built from demographic factors \cite{yang2017overcoming,chen2016neural}. Other tasks that have benefited from social representations are sarcasm detection \cite{amir2016modelling} and political opinion prediction \cite{talmacel2017predicting}.


\section{Dataset}
We experiment with the dataset of Waseem and Hovy \shortcite{c53cecce142c48628b3883d13155261c}, containing tweets manually annotated for hate speech. The authors retrieved around $136k$ tweets over a period of two months. They bootstrapped their collection process with a search for commonly used slurs and expletives related to religious, sexual, gender and ethnic minorities. From the results, they identified terms and references to entities that frequently showed up in hateful tweets. Based on this sample, they used a public Twitter \textsc{api} to collect the entire corpus of ca. $136k$ tweets.
After having manually annotated a randomly sampled subset of $16,914$ tweets under the categories \textit{racism}, \textit{sexism} or \textit{none} themselves, they asked an expert to review their annotations in order to mitigate against any biases. The inter-annotator agreement was reported at $\kappa=0.84$, with a further insight that $85\%$ of all the disagreements occurred in the \textit{sexism} class.

The dataset was released as a list of $16,907$ tweet IDs along with their corresponding annotations\footnote{\url{https://github.com/ZeerakW/hatespeech/blob/master/NAACL_SRW_2016.csv}}. Using python's \textit{Tweepy} library, we could only retrieve $16,202$ of the tweets since some of them have now been deleted or their visibility limited. Of the ones retrieved, 1,939 (12\%) are labelled as \textit{racism}, 3,148 (19.4\%) as \textit{sexism}, and the remaining 11,115 (68.6\%) as \textit{none}; this distribution follows the original dataset very closely (11.7\%, 20.0\%, 68.3\%).

We were able to extract community-based information for 1,836 out of the 1,875 unique authors who posted the $16,202$ tweets, covering a cumulative of 16,124 of them; the remaining 39 authors have either deactivated their accounts or are facing suspension. Tweets in the \textit{racism} class are from 5 of the 1,875 authors, while those in the \textit{sexism} class are from 527 of them.

\section{Methodology}
\subsection{Representing authors}
In order to leverage community-based information for the authors whose tweets form our dataset, we create an undirected unlabeled community graph wherein nodes are the authors and edges are the connections between them. An edge is instantiated between two authors $u$ and $v$ if $u$ follows $v$ on Twitter or vice versa. There are a total of 1,836 nodes and 7,561 edges. Approximately 400 of the nodes have no edges, indicating \textit{solitary} authors who neither follow any other author nor are followed by any. Other nodes have an average degree\footnote{The degree of a node is equal to the number of its direct connections to other nodes.} of 8, with close to 600 of them having a degree of at least 5. The graph is overall sparse with a density of 0.0075.

From this community graph, we obtain a vector representation, i.e., an embedding that we refer to as \textit{author profile}, for each author using the \textit{node2vec} framework \cite{node2vec-kdd2016}. \textit{Node2vec} applies the skip-gram model of Mikolov et al. \shortcite{mikolov2013efficient} to a graph in order to create a representation for each of its nodes based on their positions and their neighbors. Specifically, given a graph with nodes $V = \{v_1$, $v_2$, $\dots$, $v_n\}$, \textit{node2vec} seeks to maximize the following log probability:

\begin{equation} 
\nonumber
\sum_{v \in V} \log Pr\,(N_s(v)\, |\, v)
\end{equation}

\noindent
where $N_s(v)$ denotes the \textit{network neighborhood} of node $v$ generated through sampling strategy $s$.

In doing so, the framework learns low-dimensional embeddings for nodes in the graph. These embeddings can emphasize either their structural role or the local community they are a part of. This depends on the sampling strategies used to generate the neighborhood: if breadth-first sampling (\textsc{bfs}) is adopted, the model focuses on the immediate neighbors of a node; when depth-first sampling (\textsc{dfs}) is used, the model explores farther regions in the network, which results in embeddings that encode more information about the nodes' structural role (e.g., hub in a cluster, or peripheral node). The balance between these two ways of sampling the neighbors is directly controlled by two \textit{node2vec} parameters, namely $p$ and $q$. The default value for these is 1, which ensures a node representation that gives equal weight to both structural and community-oriented information. In our work, we use the default value for both $p$ and $q$. Additionally, since \textit{node2vec} does not produce embeddings for \textit{solitary} authors, we map these to a single zero embedding.

\begin{figure}[ht!]
\centering

\subfloat[Densely-connected authors]{
\includegraphics[width=7cm]{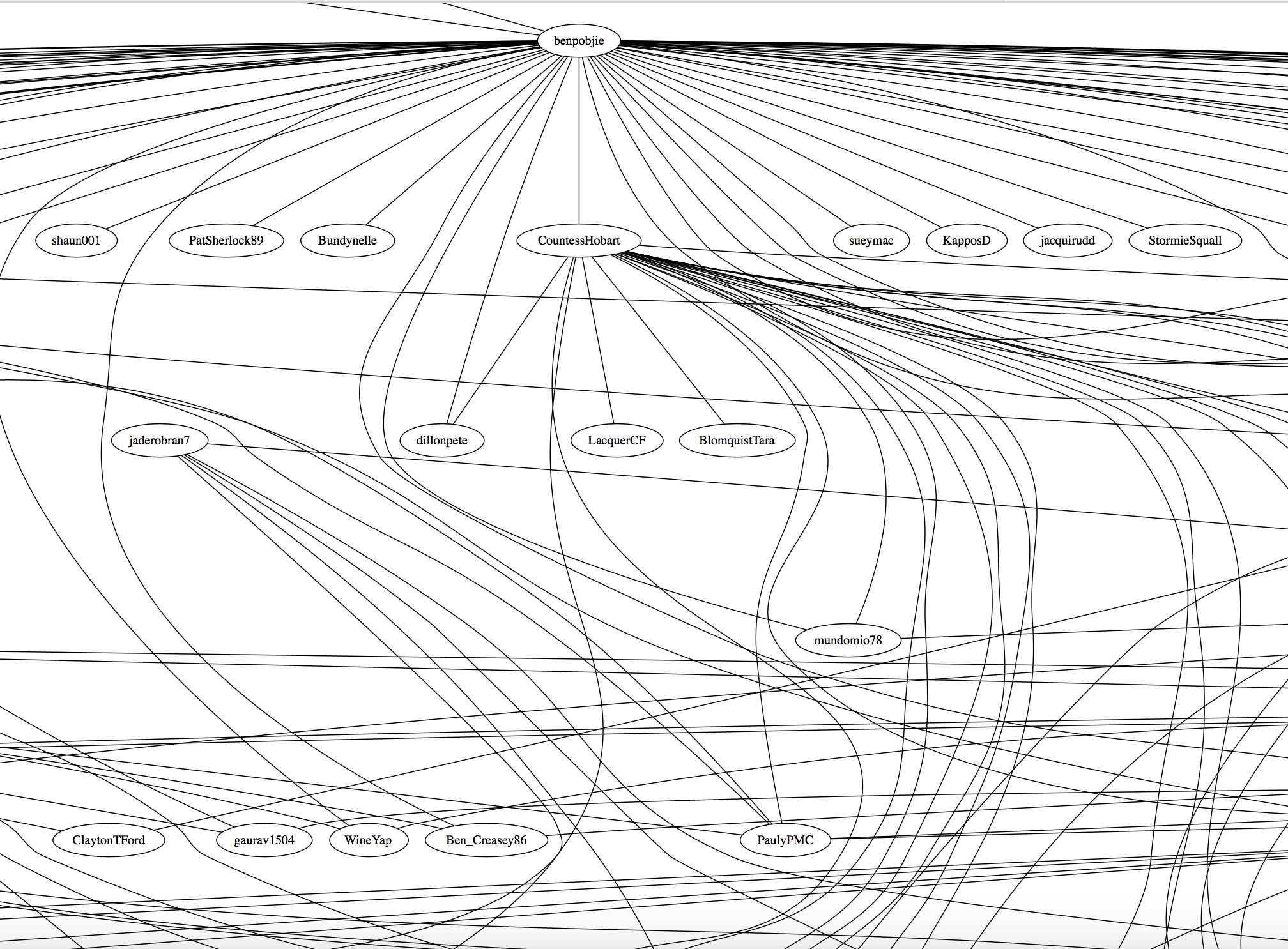}
}
\qquad
\subfloat[Sparsely-connected authors]{
\includegraphics[width=7.25cm]{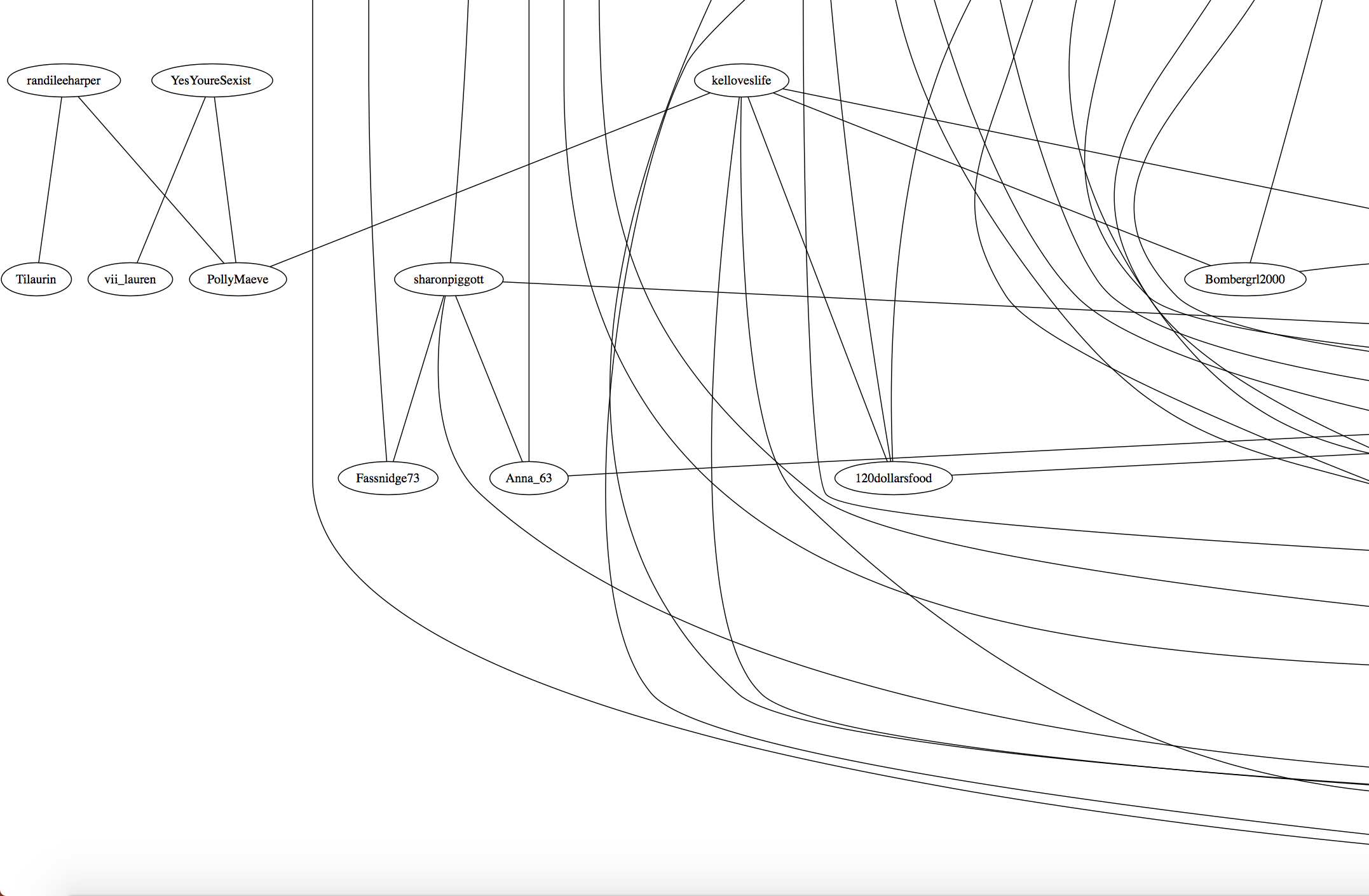}
}
\caption{Snippets from the community graph for our Twitter data.}
\label{graph}
\end{figure}

Figure \ref{graph} shows example snippets from the community graph. Some authors belong to densely-connected communities (left figure), while others are part of more sparse ones (right figure). In either case, \textit{node2vec} generates embeddings that capture the authors' neighborhood.

\subsection{Classifying content}
We experiment with seven different methods for classifying tweets as one of \textit{racism}, \textit{sexism}, or \textit{none}. We first re-implement three established and currently best-performing hate speech detection methods --- based on character n-grams and recurrent neural networks --- as our baselines. We then test whether incorporating author profiling features improves their performance.

\vspace{2 mm}
\noindent
\textbf{Char n-grams (\textsc{lr}).} As our first baseline, we adopt the method used by Waseem and Hovy \shortcite{c53cecce142c48628b3883d13155261c} wherein they train a logistic regression (\textsc{lr}) classifier on the Twitter dataset using character n-gram counts. We use uni-grams, bi-grams, tri-grams and four-grams, and \textsc{l}$_2$-normalize their counts. Character n-grams have been shown to be effective for the task of hate speech detection \cite{Nobata:2016:ALD:2872427.2883062}.

\vspace{2 mm}
\noindent
\textbf{Hidden-state (\textsc{hs}).} As our second baseline, we take the ``\textsc{rnn}'' method of Pavlopoulos et al. \shortcite{Pavlopoulos:17} which achieves state-of-the-art results on the Wikipedia datasets released by Wulczyn et al. \shortcite{Wulczyn:2017:EMP:3038912.3052591}. The method comprises a 1-layer gated recurrent unit (\textsc{gru}) that takes a sequence $w_1$, $\dots$, $w_n$ of words represented as $d$-dimensional embeddings and encodes them into hidden states $h_1$, $\dots$, $h_n$. This is followed by an \textsc{lr} layer that uses the last hidden state $h_n$ to classify the tweet. We make two minor modifications to the authors' original architecture: we deepen the 1-layer \textsc{gru} to a 2-layer \textsc{gru} and use softmax instead of sigmoid in the \textsc{lr} layer.\footnote{We also experimented with 1-layer \textsc{gru}/\textsc{lstm} and 1/2-layer bi-directional \textsc{gru}s/\textsc{lstm}s but performance only worsened or showed no gains; using sigmoid instead of softmax did not have any noteworthy effects on the results either.} Like Pavlopoulos et al., we initialize the word embeddings to \textsc{gl}o\textsc{v}e vectors \cite{Pennington:14}. In all our methods, words not available in the \textsc{gl}o\textsc{v}e set are randomly initialized in the range $\pm 0.05$, indicating the lack of semantic information. By not mapping these words to a single random embedding, we mitigate against the errors that may arise due to their conflation \cite{Madhyastha:15}. A special \textsc{oov} (out of vocabulary) token is also initialized in the same range. All the embeddings are updated during training, allowing some of the randomly-initialized ones to get task-tuned; the ones that do not get tuned lie closely clustered around the \textsc{oov} token, to which unseen words in the test set are mapped.

\vspace{2 mm}
\noindent
\textbf{Word-sum (\textsc{ws}).} As a third baseline, we adopt the ``\textsc{lstm}+\textsc{gl}o\textsc{v}e+\textsc{gbdt}" method of Badjatiya et al. \shortcite{Badjatiya:17}, which achieves state-of-the-art results on the Twitter dataset we are using. The authors first utilize an \textsc{lstm} to task-tune \textsc{gl}o\textsc{v}e-initialized word embeddings by propagating the error back from an \textsc{lr} layer. They then train a gradient boosted decision tree (\textsc{gbdt}) classifier to classify texts based on the average of the embeddings of constituent words. We make two minor modifications to this method: we use a 2-layer \textsc{gru}\footnote{We note the deeper 2-layer \textsc{gru} slightly improves performance.} instead of the \textsc{lstm} to tune the embeddings, and we train the \textsc{gbdt} classifier on the \textsc{l}$_2$-normalized sum of the embeddings instead of their average.\footnote{Although \textsc{gbdt}, as a tree based model, is not affected by the choice of monotonic function, the \textsc{l}$_2$-normalized sum ensures uniformity of range across the feature set in all our methods.} Although the authors achieved state-of-the-art results on Twitter by initializing embeddings randomly rather than with \textsc{gl}o\textsc{v}e (which is what we do here), we found the opposite when performing a 10-fold stratified cross-validation (\textsc{cv}). A possible explanation of this lies in the authors' decision to not use stratification, which for such a highly imbalanced dataset can lead to unexpected outcomes \cite{Forman:10}. Furthermore, the authors train their \textsc{lstm} on the entire dataset (including the test set) without any early stopping criterion, which leads to over-fitting of the randomly-initialized embeddings.

\vspace{2 mm}
\noindent
\textbf{Author profile (\textsc{auth}).} In order to test whether community-based information of authors is in itself sufficient to correctly classify the content produced by them, we utilize just the author profiles we generated to train a \textsc{gbdt} classifier.

\vspace{2 mm}
\noindent
\textbf{Char n-grams + author profile (\textsc{lr + auth}).} This method builds upon the \textsc{lr} baseline by appending author profile vectors on to the character n-gram count vectors for training the \textsc{lr} classifier.

\vspace{2 mm}
\noindent
\textbf{Hidden-state + author profile (\textsc{hs + auth})} and \textbf{Word-sum + author profile (\textsc{ws + auth}).} These methods are identical to the \textit{char n-grams + author profile} method 
except that here we append the author profiling features on to features derived from the \textit{hidden-state} and \textit{word-sum} baselines respectively and feed them to a \textsc{gbdt} classifier.

\section{Experiments and Results}
\subsection{Experimental setup}
We normalize the input by lowercasing all words and removing stop words. For the \textsc{gru} architecture, we use exactly the same hyper-parameters as Pavlopoulos et al. \shortcite{Pavlopoulos:17},\footnote{The authors have not released their models, and we therefore replicate their approach based on the details in their paper.} i.e., 128 hidden units, Glorot initialization, cross-entropy loss, and the Adam optimizer \cite{adam}. Badjatiya et al. \shortcite{Badjatiya:17} also use the same settings except they have fewer hidden units. In all our models, besides dropout regularization \cite{JMLR:v15:srivastava14a}, we hold out a small part of the training set as validation data to prevent over-fitting. We implement the models in \textit{Keras} \cite{keras} with \textit{Theano} back-end and use 200-dimensional pre-trained \textsc{gl}o\textsc{v}e word embeddings.\footnote{\url{http://nlp.stanford.edu/data/glove.twitter.27B.zip}} We employ \textit{Lightgbm} \cite{lightgbm} as our \textsc{gdbt} classifier and tune its hyper-parameters using 5-fold grid search. For the \textit{node2vec} framework, we use the same parameters as in the original paper \cite{node2vec-kdd2016} except we set the dimensionality of node embeddings to 200 and increase the number of iterations to 25 for better convergence.

\subsection{Results}
We perform 10-fold stratified cross validation (\textsc{cv}), as suggested by Forman and Scholz \shortcite{Forman:10}, to evaluate all seven methods described in the previous section. Following previous research \cite{Badjatiya:17,W17-3006}, we report the average weighted precision, recall, and \textsc{f}$_1$ scores for all the methods. The average weighted precision is calculated as:

\begin{equation}
\nonumber
\frac{\sum_{i=1}^{10}\; (w_r\cdot\textrm{P}_r^i + w_s\cdot\textrm{P}_s^i + w_n\cdot\textrm{P}_n^i)}{10}
\end{equation}

\noindent
where $\textrm{P}_r^i, \textrm{P}_s^i, \textrm{P}_n^i$ are precision scores on the \textit{racism}, \textit{sexism}, and \textit{none} classes from the $i^{th}$ fold of the \textsc{cv}. The values $w_r$, $w_s$, and $w_n$ are the proportions of the \textit{racism}, \textit{sexism}, and \textit{none} classes in the dataset respectively; since we use stratification, these proportions are constant ($w_r=0.12$, $w_s=0.19$, $w_n=0.69$) across all folds. Average weighted recall and \textsc{f}$_1$ are calculated in the same manner.

The results are presented in Table \ref{twitter}. For all three baseline methods (\textsc{lr}, \textsc{ws}, and \textsc{hs}), the addition of author profiling features significantly improves performance ($p < 0.05$ under 10-fold \textsc{cv} paired t-test). The \textsc{lr + auth} method yields the highest performance of \textsc{f}$_1$ $=87.57$, exceeding its respective baseline by nearly 4 points. A similar trend can be observed for the other methods as well. These results point to the importance of community-based information and author profiling in hate speech detection and demonstrate that our approach can further improve the performance of existing state-of-the-art methods.

\begin{table}[ht!]
\centering
\small
\begin{tabular}{| c | c | c | c | c |}
\hline
& \textbf{Method} & \textbf{\textsc{p}} & \textbf{\textsc{r}} & \textbf{\textsc{f}$_1$}\\ \hline
\multirow{3}{*}{Baselines}& \textsc{lr} & 84.07 & 84.31 & 83.81\\
& \textsc{hs} & 83.50 & 83.71 & 83.54\\
& \textsc{ws} & 82.86 & 83.10 & 82.37\\ \hline
\multirow{4}{*}{Our methods} & \textsc{auth} & 72.13 & 76.05 & 71.26\\
& \textsc{lr + auth} & \textbf{87.57} & \textbf{87.66} & \textbf{87.57}\\
& \textsc{hs + auth} & 87.29 & 87.32 & 87.29\\
& \textsc{ws + auth} & 87.11 & 87.20 & 87.08\\\hline
\end{tabular}
\caption{Average weighted precision, recall and \textsc{f}$_1$ scores of the different methods on the Twitter datasest. All improvements are significant ($p < 0.05$) under 10-fold \textsc{cv} paired t-test.}
\label{twitter}
\end{table}

\begin{table}[ht!]
\centering
\small
\subfloat[\textit{Racism} class]{\begin{tabular}{| c | c | c | c |}
\hline
\textbf{Method} & \textbf{\textsc{p}} & \textbf{\textsc{r}} & \textbf{\textsc{f}$_1$}\\ \hline
\textsc{lr} & \textbf{77.29} & 67.92 & 72.28\\
\textsc{hs} & 74.15 & 72.46 & 73.24\\
\textsc{ws} & 76.43 & 67.77 & 71.78\\ \hline
\textsc{auth} & 43.33 & 0.31 & 0.61\\
\textsc{lr + auth} & 76.10 & \textbf{74.16} & \textbf{75.09}\\
\textsc{hs + auth} & 74.42 & 73.54 & 73.91\\
\textsc{ws + auth} & 75.12 & 72.46 & 73.72\\ \hline
\end{tabular}}
\qquad
\subfloat[\textit{Sexism} class]{\begin{tabular}{| c | c | c | c |}
\hline
\textbf{Method} & \textbf{\textsc{p}} & \textbf{\textsc{r}} & \textbf{\textsc{f}$_1$}\\ \hline
\textsc{lr} & 82.66 & 63.98 & 72.09\\
\textsc{hs} & 76.04 & 68.84 & 72.24\\
\textsc{ws} & 81.75 & 57.37 & 67.38\\ \hline
\textsc{auth} & 66.85 & 75.44 & 70.88\\
\textsc{lr + auth} & 86.22 & 79.07 & 82.47\\
\textsc{hs + auth} & 84.15 & \textbf{81.32} & \textbf{82.75}\\
\textsc{ws + auth} & \textbf{86.37} & 77.92 & 81.91\\ \hline
\end{tabular}}
\caption{Performance of the methods on the \textit{racism} and \textit{sexism} classes separately. All improvements are significant ($p < 0.05$) under 10-fold \textsc{cv} paired t-test.}
\label{twitter_neg}
\end{table}

In Table \ref{twitter_neg}, we further compare the performance of the different methods on the \textit{racism} and \textit{sexism} classes individually. As in the previous experiments, the scores are averaged over 10 folds of \textsc{cv}. Of particular interest are the scores for the \textit{sexism} class where the \textsc{f}$_1$ increases by over 10 points upon the addition of author profiling features. Upon analysis, we find that such a substantial increase in performance stems from the fact that many of the 527 unique authors of the sexist tweets are closely connected in the community graph. This allows for their penchant for sexism to be expressed in their respective author profiles.

The author profiling features on their own (\textsc{auth}) achieve impressive results overall and in particular on the \textit{sexism} class, where their performance is typical of a community-based generalization, i.e., low precision but high recall. For the \textit{racism} class on the other hand, the performance of \textsc{auth} on its own is quite poor. This contrast can be explained by the fact that tweets in the \textit{racism} class come from only 5 unique authors who: (i) are isolated in the community graph, or (ii) have also authored several tweets in the \textit{sexism} class, or (iii) are densely connected to authors from the \textit{sexism} and \textit{none} classes which possibly camouflages their racist nature.

We believe that the gains in performance will be more pronounced as the underlying community graph grows since there will be less solitary authors and more edges worth harnessing information from.\footnote{Regarding the scalability of our approach, we quote the authors of \textit{node2vec}: ``The major phases of \textit{node2vec} are trivially parallelizable, and it can scale to large networks with millions of nodes in a few hours''. } Even when the data is skewed and there is an imbalance of hateful vs. non-hateful authors, we do expect our approach to still be able to identify clusters of authors with similar views.

\section{Analysis and discussion}
We conduct a qualitative analysis of system errors and the cases where author profiling leads to the correct classification of previously misclassified examples. Table \ref{win_examples} shows examples of hateful tweets from the dataset that are misclassified by the \textsc{lr} method, but are correctly classified upon the addition of author profiling features, i.e., by the \textsc{lr + auth} method. It is worth noting that some of the wins scored by the latter are on tweets that are part of a larger hateful discourse or contain links to hateful content while not explicitly having textual cues that are indicative of hate speech per se. The addition of author profiling features may then be viewed as a proxy for wider discourse information, thus allowing us to correctly resolve the cases where lexical and semantic features alone are insufficient.\footnote{We note that the annotators of the dataset took discourse into account when annotating the tweets. However, the dataset was released as a list of tweet \textsc{id} and corresponding annotation (racism/sexism/none) pairs; there is no annotation available regarding which tweets are related to which other ones.}

\begin{table}[ht!]
\centering
\small
\begin{tabular}{| p{8cm} | c | c |}
\hline
\hspace{3.6cm}\textbf{Tweet} & \multicolumn{2}{c |}{\textbf{Predicted label}} \\ \hline
 & \textbf{\textsc{lr}} & \textbf{\textsc{lr + auth}} \\ \hline
\textit{@Mich\_McConnell Just ``her body" right?} & \textit{none} & \textit{sexism}\\
\textit{@Starius: \#GamerGate https://t.co/xuFwsIgxFK WE WIN! ahahahaha} & \textit{none} & \textit{sexism}\\
\textit{\#Islam dominates our crime, prison \& welfare system \& national security. Why are we still importing it? @PeterDutton\_MP  \#amagenda \#auspol} & \textit{none} & \textit{racism}\\
\textit{@Wateronatrain: @MT8\_9 You might like this \#patriarchy http://t.co/c9m2pFmFJ3} & \textit{none} & \textit{sexism}\\
\textit{It seems that Allah sits around all day obsessing about women's hands and faces showing. I guess idiots need a god on their level. \#Islam} & \textit{none} & \textit{racism}\\
\textit{@SalemP08: @MT8\_9 @LiljaOB @midnitebacon @Superjutah @Transic\_nyc her response is pretty terrifying.} & \textit{none} & \textit{sexism}\\
\textit{@JosephIsVegan @SumbelinaZ @IronmanL1 @Hatewatch Why would you profile white people. Blacks murder at 6 times the rate as whites.} & \textit{none} & \textit{racism}\\ \hline
\end{tabular}
\caption{Examples of improved classification upon the addition of author profiling features (\textsc{auth}).}
\label{win_examples}
\end{table}

However, a number of hateful tweets still remain misclassified despite the addition of author profiling features. According to our analysis, many of these tend to contain \textsc{url}s to hateful content, e.g., ``\textit{@salmonfarmer1: Logic in the world of Islam http://t.co/6nALv2HPc3}" and ``\textit{@juliarforster Yes. http://t.co/ixbt0uc7HN}". Since Twitter shortens all \textsc{url}s into a standard format, there is no indication of what they refer to. One way to deal with this limitation could be to additionally maintain a blacklist of links.
Another source of system errors is the deliberate obfuscation of words by authors in order to evade detection, e.g., ``\textit{Kat, a massive c*nt. The biggest ever on \#mkr \#cuntandandre}". Current hate speech detection methods, including ours, do not directly attempt to address this issue.
While this is a challenge for bag-of-word based methods such as \textsc{lr}, we hypothesize that neural networks operating at the character level may be helpful in recognizing obfuscated words.

\begin{figure}[ht!]
\centering
\includegraphics[width=7cm]{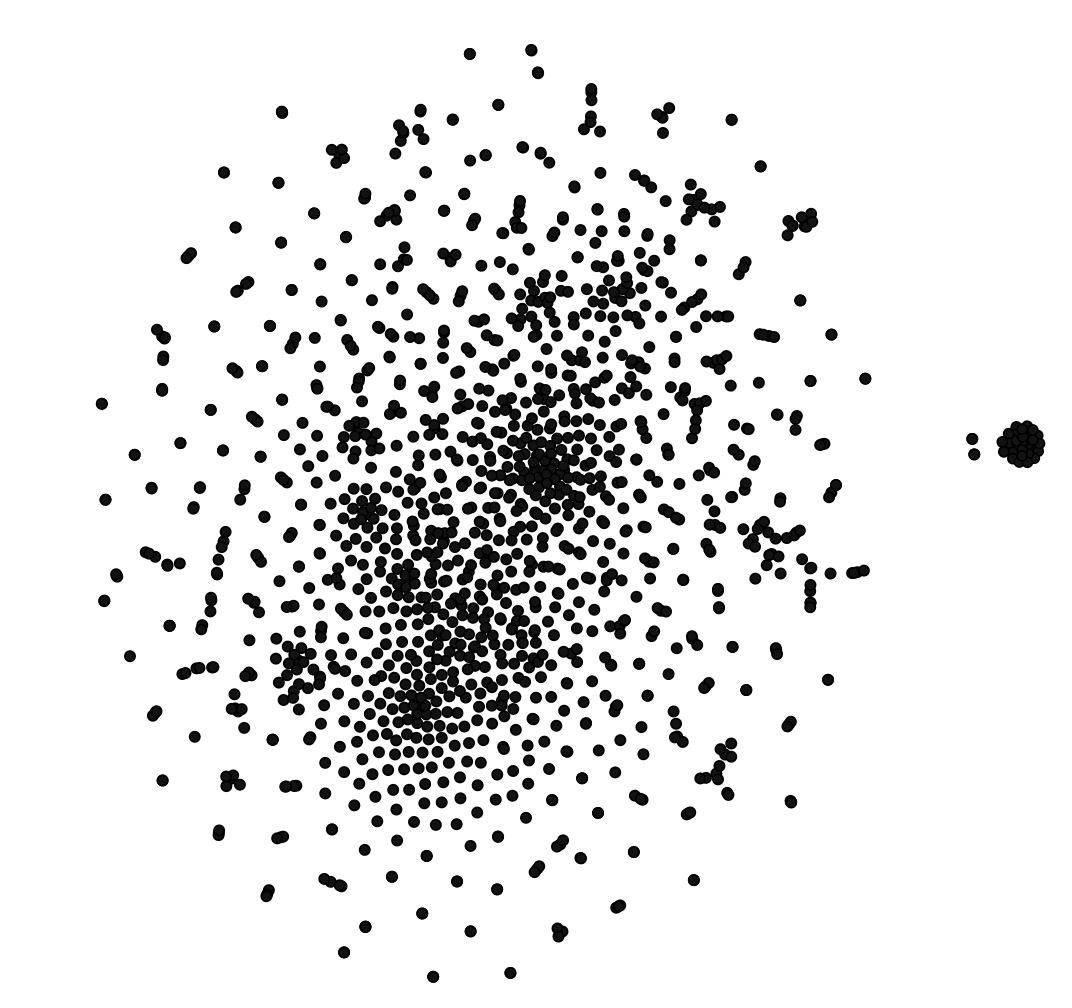}
\caption{Visualization of author embeddings in 2-dimensional space.}
\label{tsne}
\end{figure}

\begin{figure}[ht!]
\centering

\subfloat[\textit{None} class]{
\includegraphics[width=7cm]{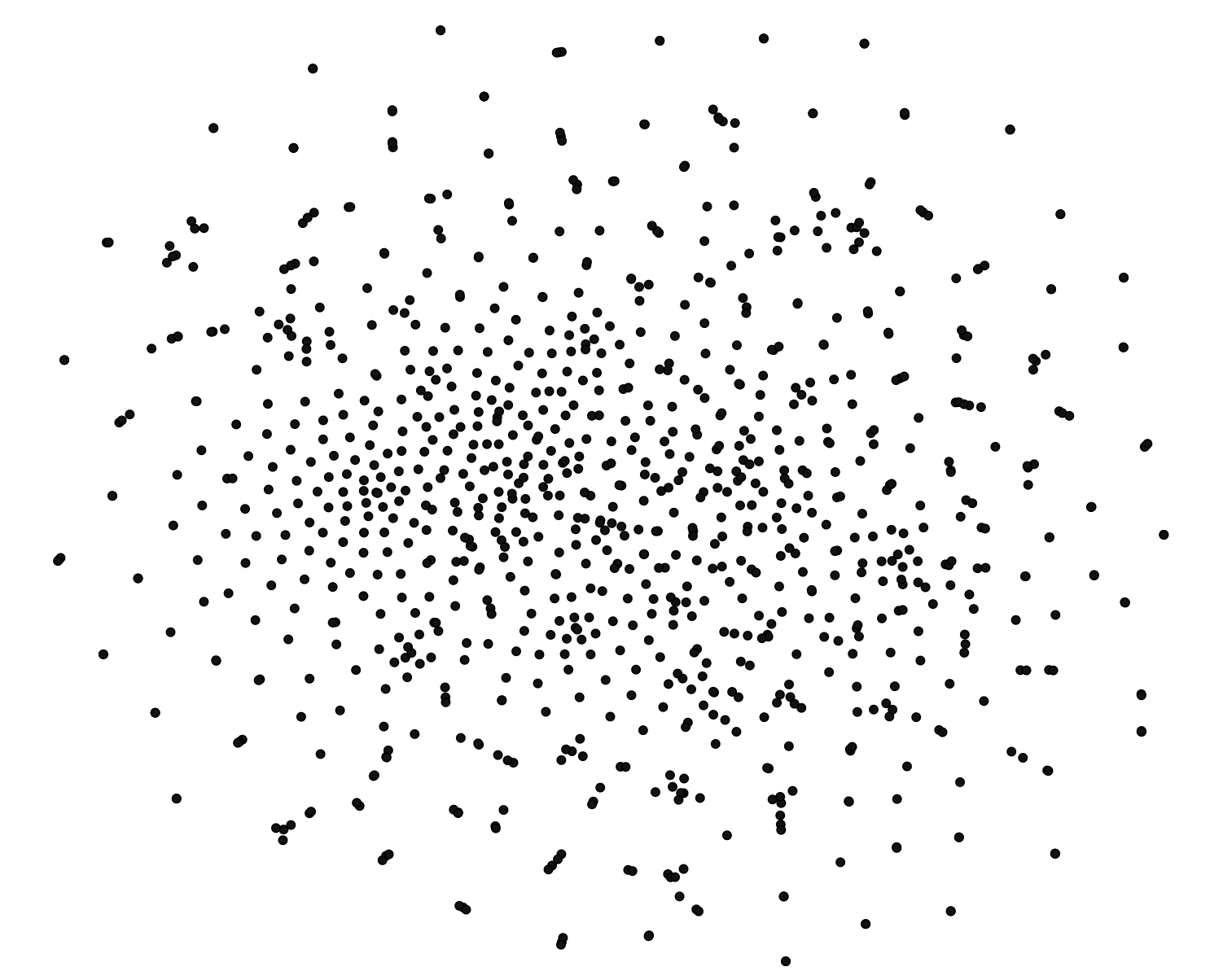}
}
\qquad
\subfloat[\textit{Sexism} class]{
\includegraphics[width=7cm]{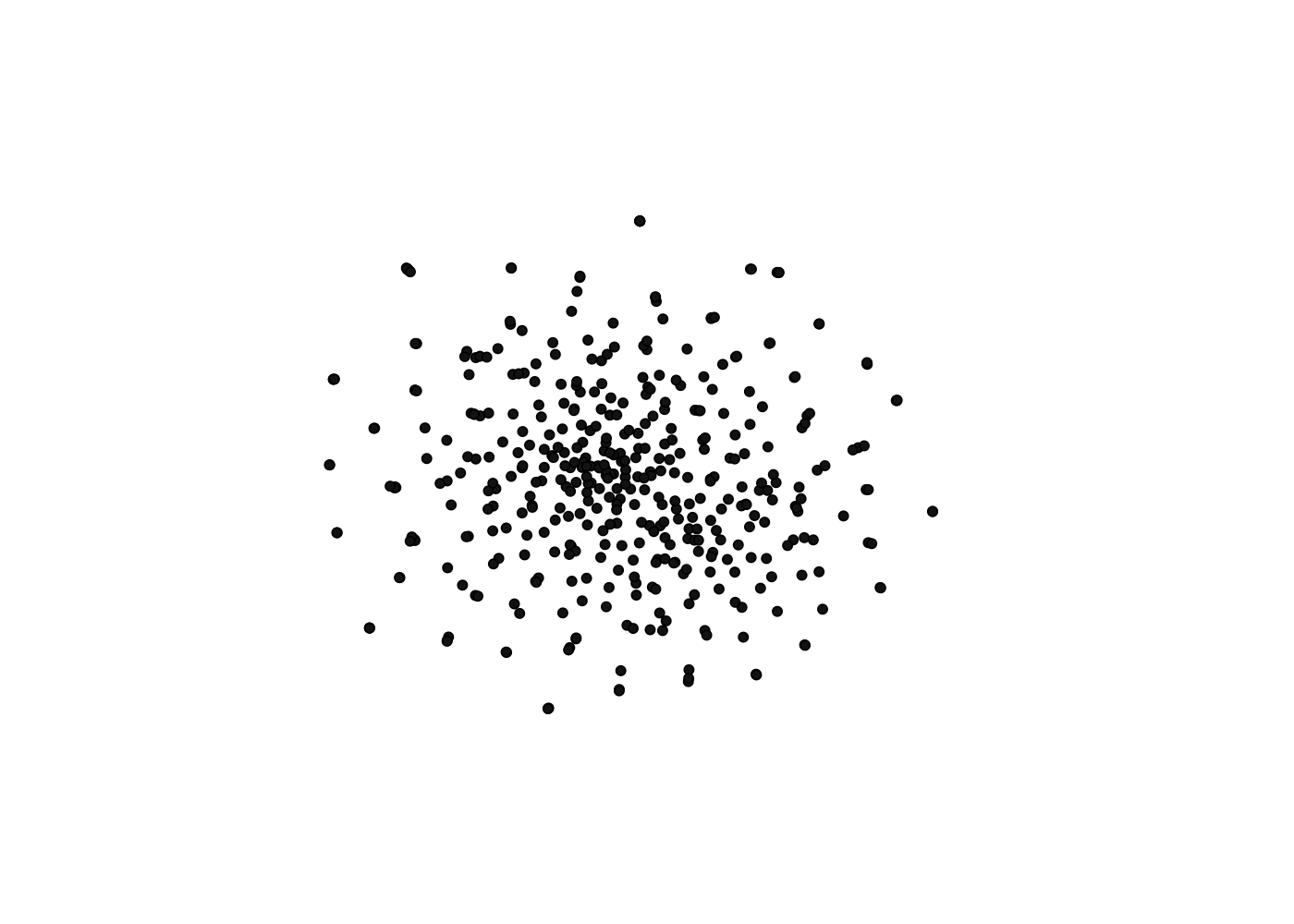}
}
\caption{Visualization of authors from different classes.}
\label{tnse_class}
\end{figure}

We further conducted an analysis of the author embeddings generated by \textit{node2vec}, in order to validate that they capture the relevant aspects of the community graph. We visualized the author embeddings in 2-dimensional space using \textit{t}-\textsc{sne} \cite{tsne}, as shown in Figure \ref{tsne}. We observe that, as in the community graph, there are a few densely populated regions in the visualization that represent authors in closely knit groups who exhibit similar characteristics. The other regions are largely sparse with smaller clusters. Note that we exclude \textit{solitary} users from this visualization since we have to use a single zero embedding to represent them.

Figure \ref{tnse_class} further provides visualizations for authors from the \textit{sexism} and \textit{none} classes separately. While the authors from the \textit{none} class are spread out in the embedding space, the ones from the \textit{sexism} class are more tightly clustered. Note that we do not visualize the 5 authors from the \textit{racism} class since 4 of them are already covered in the \textit{sexism} class.

\section{Conclusions}
In this paper, we explored the effectiveness of community-based information about authors for the purpose of identifying hate speech. Working with a dataset of $16k$ tweets annotated for \textit{racism} and \textit{sexism}, we first comprehensively replicated three established and currently best-performing hate speech detection methods based on character n-grams and recurrent neural networks as our baselines. We then constructed a graph of all the authors of tweets in our dataset and extracted community-based information in the form of dense low-dimensional embeddings for each of them using \textit{node2vec}. We showed that the inclusion of author embeddings significantly improves system performance over the baselines and advances the state of the art in this task. Users prone to hate speech do tend to form social groups online, and this stresses the importance of utilizing community-based information for automatic hate speech detection. In the future, we wish to explore the effectiveness of community-based author profiling in other tasks such as stereotype identification and metaphor detection.

\bibliography{coling2018}
\bibliographystyle{coling2018}
\end{document}